\documentclass[journal]{IEEEtran}
\usepackage{cite}
\usepackage{bm}
\usepackage{flushend}
\usepackage{stfloats}
\usepackage{multicol}
\usepackage{multirow}
\usepackage{array}
\usepackage{graphicx}
\usepackage{epstopdf}
\usepackage{setspace}
\usepackage{lettrine}
\usepackage{amsmath}
\usepackage{amsbsy}
\usepackage{mathrsfs}
\usepackage{amsfonts,amssymb}
\usepackage{algorithm}
\usepackage{algorithmic}
\renewcommand{\algorithmicrequire}{\textbf{Input:}} % Use Input in the format of Algorithm
\renewcommand{\algorithmicensure}{\textbf{Output:}} % Use Output in the format of Algorithm
\usepackage{graphicx}
\usepackage{subfigure}
\usepackage{booktabs}
\usepackage{amsmath} % assumes amsmath package installed
\usepackage{amssymb}  % assumes amsmath package installed

\begin{document}

%\title{ \LARGE{Kernel Risk-Sensitive Loss: Definition, Properties and Application to Robust Adaptive Filtering}}
\title{ \huge{Quantized Minimum Error Entropy Criterion}}

\author{Badong Chen, \emph{Senior Member, IEEE}, Lei Xing, \emph{Student Member, IEEE},\\
	 Nanning Zheng, \emph{Fellow, IEEE}, Jos\'e C. Pr\'incipe, \emph{Fellow IEEE}}

\maketitle

\begin{abstract}
Comparing with traditional learning criteria, such as mean square error (MSE), the \textit{minimum error entropy} (MEE) criterion is superior in nonlinear and non-Gaussian signal processing and machine learning. The argument of the logarithm in Renyi’s entropy estimator, called \textit{information potential} (IP), is a popular MEE cost in \textit{information theoretic learning} (ITL). The computational complexity of IP is however quadratic in terms of sample number due to double summation. This creates computational bottlenecks especially for large-scale datasets. To address this problem, in this work we propose an efficient quantization approach to reduce the computational burden of IP, which decreases the complexity from $O\left( {{N^2}} \right)$ to $O\left( {{MN}} \right)$  with $M \ll N$. The new learning criterion is called the \textit{quantized MEE} (QMEE). Some basic properties of QMEE are presented. Illustrative examples are provided to verify the excellent performance of QMEE.

\end{abstract}


\let\thefootnote\relax\footnotetext{This work was supported by 973 Program (No. 2015CB351703) and National NSF of China (No. 91648208, No. 61372152, No. U1613219).
\par Badong Chen, Lei Xing, Nanning Zheng, and Jos\'e C. Pr\'incipe are with the Institute of Artificial Intelligence and Robotics, Xi'an Jiaotong University, Xi'an, China. ({chenbd; nnzheng }@ mail.xjtu.edu.cn; xl2010@stu.xjtu.edu.cn; principe@cnel.ufl.edu), Jos\'e C. Pr\'incipe is also with the Department of Electrical and Computer Engineering, University of Florida, Gainesville, USA.
}


% \fi
\IEEEpeerreviewmaketitle
\section{Introduction}
\lettrine[lines=2]{A}{S} a well-known learning criterion in \textit{information theoretic learning} (ITL) \cite{principe2010information,chen2013system,erdogmus2006linear}, the \textit{minimum error entropy} (MEE) finds successful applications in various learning tasks, including regression, classification, clustering, feature selection and many others \cite{erdogmus2002error,erdogmus2002generalized,silva2006error,de2013minimum,gokcay2002information,santamaria2002entropy,han2007minimum,bessa2009entropy,wang2015minimum,chen2010mean,chen2012survival,chen2013kernel,wu2015minimum,shen2015minimum}. The basic principle of MEE is to learn a model to discover structure in data by minimizing the entropy of error between model and data generating system \cite{principe2010information}. Entropy takes all higher order moments into account and hence, is a global descriptor of the underlying distribution. The MEE can perform much better than the traditional mean square error (MSE) criterion that considers only the second order moment of the error, especially in nonlinear and non-Gaussian (multi-peak, heavy-tailed, etc.) signal processing and machine learning.
\par In practical applications, an MEE cost can be estimated based on a PDF estimator. The most widely used MEE cost in ITL is the \textit{information potential} (IP), which is the argument of the logarithm in Renyi’s entropy \cite{principe2010information}. The IP can be estimated directly from data and computed by a double summation over all samples. This is much different from traditional learning costs that only involve a single summation.  Although IP is simpler than many other entropic costs, it is still computationally very expensive due to the pairwise computation (i.e. double summation). This may pose computational bottlenecks for large-scale datasets. To address this issue, we propose in this paper an efficient approach to decrease the computational complexity of IP from $O\left( {{N^2}} \right)$ to $O\left( {{MN}} \right)$  with $M \ll N$. The basic idea is to simplify the inner summation by quantizing the error samples with a simple quantization method. The simplified learning criterion is called the \textit{quantized MEE} (QMEE). Some properties of the QMEE are presented, and the desirable performance of QMEE is confirmed by several illustrative results.
\par The remainder of the paper is organized as follows. The MEE criterion is briefly reviewed in section II. The QMEE is proposed in section III. The illustrative examples are provided in section IV and finally, the conclusion is given in section V.

\section{BRIEF REVIEW OF MEE CRITERION}
Consider learning from $N$ examples ${Z^N} = \left\{ {\left( {{x_i},{y_i}} \right) \in \pmb{\mathcal{X}} \times \pmb{\mathcal{Y}}} \right\}$, $i = 1,2, \cdots ,N$ , which are drawn independently from an unknown probability distribution $\pmb{\mathcal{D}}$ on $\mathcal{Z}: = \pmb{\mathcal{X}} \times \pmb{\mathcal{Y}}$ . Here we assume $\pmb{\mathcal{X}} \subset \mathbb{R}^{d}$ and $\pmb{\mathcal{Y}} \subset \mathbb{R}$. Usually, a loss function $\ell \left( {f,(x,y)} \right)$ is used to measure the performance of the hypothesis $f:\pmb{\mathcal{X}} \to \pmb{\mathcal{Y}}$. For regression, one can choose the squared error loss $\ell \left( {f,(x,y)} \right) = {\left( {y - f(x)} \right)^2} = {e^2}$, where $e = y - f(x) \in \mathbb{R} $ is the prediction error. Then the goal of learning is to find a solution in hypothesis space that minimizes the expected cost function $\textbf{E}\left[ {\ell \left( {f,(x,y)} \right)} \right]$, where the expectation is taken over $\pmb{\mathcal{D}}$. As the distribution $\pmb{\mathcal{D}}$ is unknown, in general we use the empirical cost function:

\begin{equation}\label{cost1}
J = \frac{1}{N}\sum\limits_{i = 1}^N {\ell \left( {f,({x_i},{y_i})} \right)} 
\end{equation}

\noindent which involves a summation over all samples. Sometimes, a regularization term is added to the above sum to prevent overfitting. Under MSE criterion, the empirical cost function becomes
\begin{equation}\label{cost2}
{J_{MSE}} = \frac{1}{N}\sum\limits_{i = 1}^N {{{\left( {{y_i} - f({x_i})} \right)}^2}}  = \frac{1}{N}\sum\limits_{i = 1}^N {e_i^2} 
\end{equation}

\noindent where ${e_i} = {y_i} - f({x_i})$ is the prediction error for sample $\left( {{x_i},{y_i}} \right)$. The computational complexity for evaluating the above cost and its gradient with respect to ${e_i}$ ($i = 1,2, \cdots ,N$ ) is $O(N)$.

\par In the context of \textit{information theoretic learning} (ITL), one can adopt Renyi’s entropy of order $\alpha$ ($\alpha  > 0$, $\alpha  \ne 1$) as the cost function \cite{principe2010information}:
\begin{equation}\label{Renyi}
{H_\alpha }(e) = \frac{1}{{1 - \alpha }}\log \int {{p^\alpha }(e)de} 
\end{equation}

\noindent where $p(.)$ denotes the error’s PDF. Under MEE criterion, the optimal hypothesis can thus be solved by minimizing the error entropy ${H_\alpha }(e)$. The argument of the logarithm in ${H_\alpha }(e)$, called \textit{information potential} (IP), is 
\begin{equation}\label{IP}
{I_\alpha }(e) = \int {{p^\alpha }(e)de}  = \textbf{E}\left[ {{p^{\alpha  - 1}}(e)} \right]
\end{equation}

\noindent Since the logarithm function is a monotonically increasing function, minimizing Renyi’s entropy ${H_\alpha }(e)$ is equivalent to minimizing (for $\alpha  < 1$) or maximizing (for $\alpha  > 1$) the IP ${I_\alpha }(e)$. In ITL, for simplicity the parameter $\alpha$ is usually set at $\alpha  = 2$.  In the rest of the paper, without loss of generality we only consider the case of $\alpha  = 2$. In this case, we have
\begin{equation}\label{renyi2}
\min {H_2}(e) \Leftrightarrow \max {I_2}(e) = \textbf{E}\left[ {p(e)} \right]
\end{equation}

\noindent According to ITL \cite{principe2010information}, an empirical version of the quadratic IP can be expressed as
\begin{equation}\label{e_ip}
{\hat I_2}(e){\rm{ = }}\frac{1}{N}\sum\limits_{i = 1}^N {\hat p\left( {{e_i}} \right)}  = \frac{1}{{{N^2}}}\sum\limits_{i = 1}^N {\sum\limits_{j = 1}^N {{G_\sigma }\left( {{e_i} - {e_j}} \right)} } 
\end{equation}

\noindent where $\hat p(.)$ is Parzen's PDF estimator \cite{silverman1986density}:
\begin{equation}\label{p_ip}
\hat p\left( x \right) = \frac{1}{N}\sum\limits_{j = 1}^N {{G_\sigma }\left( {x - {e_j}} \right)} 
\end{equation}

\noindent with ${G_\sigma }(.)$ being the Gaussian kernel with bandwidth $\sigma$:
\begin{equation}\label{gaussian_kernel}
{G_\sigma }(x) = \frac{1}{{\sqrt {2\pi } \sigma }}\exp \left( { - \frac{{{x^2}}}{{2{\sigma ^2}}}} \right)
\end{equation}

\noindent The PDF estimator $\hat p(.)$ can be viewed as an \textit{adaptive loss} function that varies with the error samples $\left\{ {{e_1},{e_2}, \cdots ,{e_N}} \right\}$. This is much different from the conventional loss functions that are typically left unchanged after being set. For example, the loss function of MSE is always $\ell (x) = {x^2}$. The adaptation of loss function is potentially beneficial because the risk is matched to the error distribution. The superior performance of MEE has been shown theoretically as well as confirmed numerically \cite{principe2010information}. However, the price we have to pay is that there is a \textit{double summation} over all samples, which is obviously time consuming especially for large-scale datasets. The computational complexity for evaluating the cost function (\ref{e_ip}) is $O(N^2)$. The goal of this work is to find an efficient way to simplify the computation of the empirical IP.

\section{QUANTIZED MEE}
Comparing with conventional cost functions for machine learning, the MEE cost (or equivalently, the IP) involves an additional summation operation, namely the computation of the PDF estimator. The basic idea of our approach is thus to reduce the computational burden of the PDF estimation (i.e. the inner summation). We aim to estimate the error’s PDF from fewer samples. A natural way is to represent the $N$ error samples $\left\{ {{e_1},{e_2}, \cdots ,{e_N}} \right\}$ with a smaller data set by using a simple quantization method. Of course, the quantization will decrease the accuracy of PDF estimation. However, the PDF estimator for an entropic cost function is very different from the ones for traditional density estimation. Indeed, for a cost function for machine learning, ultimately what’s going to matter is the extrema (maxima or minima) of the cost function, not the exact value of the cost. Our experimental results have shown that with quantization the MEE can achieve almost the same (or even better) performance as the original MEE learning.

Let $Q[.]$ denote a quantization operator (or quantizer) with a codebook $C$ containing $M$ (in general $M \ll N$) real valued code words, i.e. $C = \left\{ {{c_1},{c_2}, \cdots ,{c_M} \in \mathbb{R}} \right\}$. Then $Q[.]$ is a function that can map the error sample ${e_j}$ into one of the $M$ code words in $C$, i.e. $Q[{e_j}] \in C$ . In this work, we assume that each error sample is quantized to the nearest code word. With the quantizer $Q[.]$, the empirical IP in (\ref{e_ip}) can be simplified to
\begin{equation}\label{simple_eip}
\begin{aligned}
{{\hat I}_2}(e)&{\rm{ = }}\frac{1}{{{N^2}}}\sum\limits_{i = 1}^N {\sum\limits_{j = 1}^N {{G_\sigma }\left( {{e_i} - {e_j}} \right)} } \\
&{\rm{       }} \approx \hat I_2^Q(e) = \frac{1}{{{N^2}}}\sum\limits_{i = 1}^N {\sum\limits_{j = 1}^N {{G_\sigma }\left( {{e_i} - Q[{e_j}]} \right)} } \\
&{\rm{       }} = \frac{1}{{{N^2}}}\sum\limits_{i = 1}^N {\left( {\sum\limits_{m = 1}^M {{M_m}{G_\sigma }\left( {{e_i} - {c_m}} \right)} } \right)} \\
&{\rm{       }} = \frac{1}{N}\sum\limits_{i = 1}^N {{{\hat p}_Q}\left( {{e_i}} \right)} 
\end{aligned}
\end{equation}

\noindent where ${M_m}$ is the number of error samples that are quantized to the code word ${c_m}$, and ${\hat p_Q}\left( x \right){\rm{ = }}\frac{1}{N}\sum\limits_{m = 1}^M {{M_m}{G_\sigma }\left( {x - {c_m}} \right)} $ is the PDF estimator based on the quantized error samples. Clearly, we have $\sum\limits_{m = 1}^M {{M_m}}  = N$ and $\int {{{\hat p}_Q}\left( x \right)dx} {\rm{ = 1}}$.\\
\textit{Remark}: The computational complexity of the \textit{quantized MEE} (QMEE) cost $\hat I_2^Q(e)$ is $O\left( {MN} \right)$, which is much simpler than the original cost of (\ref{e_ip}) especially for large-scale datasets ($M \ll N$ ).\\
\par Before designing the quantizer $Q[.]$, we present below some basic properties of the QMEE cost. 

\noindent \textit{Property 1}: When the codebook $C{\rm{ = }}\left\{ {{e_1},{e_2}, \cdots ,{e_N}} \right\}$, we have $\hat I_2^Q(e){\rm{ = }}{\hat I_2}(e)$.

\noindent \textit{Proof}: Straightforward since in this case we have $Q[{e_j}] = {e_j}$, $j = 1,2, \cdots ,N$ .

\noindent \textit{Property 2}: The QMEE cost $\hat I_2^Q(e)$ is bounded, i.e. $\hat I_2^Q(e) \le \frac{1}{{\sqrt {2\pi } \sigma }}$, with equality if and only if ${e_1} = {e_2} =  \cdots  = {e_M} = c$, where $c$ is an element of $C$.

\noindent \textit{Proof}: Since ${G_\sigma }(x) \le \frac{1}{{\sqrt {2\pi } \sigma }}$ with equality if and only if $x = 0$, we have

\begin{equation}\label{p2}
\begin{aligned}
\hat I_2^Q(e) &= \frac{1}{{{N^2}}}\sum\limits_{i = 1}^N {\sum\limits_{j = 1}^N {{G_\sigma }\left( {{e_i} - Q[{e_j}]} \right)} }\\
&\le \frac{1}{{{N^2}}}\sum\limits_{i = 1}^N {\sum\limits_{j = 1}^N {\frac{1}{{\sqrt {2\pi } \sigma }}} }  = \frac{1}{{\sqrt {2\pi } \sigma }}
\end{aligned}
\end{equation}

\noindent with equality if and only if ${e_i} = Q[{e_j}]$ , $\forall i,j$ , which means ${e_1} = {e_2} =  \cdots  = {e_M} = c$ .

\noindent \textit{Property 3}: It holds that $\hat I_2^Q(e) = \frac{1}{M}\sum\limits_{j = 1}^M {{\alpha _m}\hat p\left( {{c_m}} \right)} $, where ${\alpha _m}{\rm{ = }}\frac{{{M_m}}}{N}$,  satisfying $\sum\limits_{m = 1}^M {{\alpha _m}} {\rm{ = }}1$.

\noindent \textit{Proof}: One can easily derive

\begin{equation}\label{p3}
\begin{aligned}
\hat I_2^Q(e) &= \frac{1}{{{N^2}}}\sum\limits_{i = 1}^N {\left( {\sum\limits_{m = 1}^M {{M_m}{G_\sigma }\left( {{e_i} - {c_m}} \right)} } \right)} \\
&{\rm{       }} = \sum\limits_{m = 1}^M {\frac{{{M_m}}}{N}\left( {\frac{1}{N}\sum\limits_{i = 1}^N {{G_\sigma }\left( {{e_i} - {c_m}} \right)} } \right)} \\
&{\rm{       }} = \sum\limits_{m = 1}^M {{\alpha _m}\hat p\left( {{c_m}} \right)} 
\end{aligned}
\end{equation}

\textit{Remark}: By Property 3, the QMEE cost $\hat I_2^Q(e)$ is equal to a weighted average of the Parzen's PDF estimator evaluated at the code words. Moreover, when there is only one code word in $C$, i.e. $C{\rm{ = }}\left\{ c \right\}$, we have $\hat I_2^Q(e) = \hat p\left( c \right)$. In particular, when $C{\rm{ = }}\left\{ 0 \right\}$, we have $\hat I_2^Q(e) = \hat V(e) = \frac{1}{N}\sum\limits_{i = 1}^N {{G_\sigma }\left( {{e_i}} \right)}  = \hat p\left( 0 \right)$, where $\hat V(e)$ denotes the empirical correntropy \cite{liu2007correntropy,he2011robust,chen2012maximum,chen2015convergence,zhao2011kernel}, which is a well-known local similarity measure in ITL. In this sense, the correntropy can be viewed as a special case of the QMEE cost. Actually, the correntropy measures the local similarity about the zero, while QMEE cost $\hat I_2^Q(e)$ measures the average similarity about every code word in $C$.

\noindent \textit{Property 4}: When $\sigma$ is large enough, we have $\hat I_2^Q(e) \approx \frac{1}{{\sqrt {2\pi } \sigma }} - \frac{1}{{2\sqrt {2\pi } {\sigma ^3}}}\sum\limits_{m = 1}^M {{\alpha _m}{\mu _m}} $, where ${\mu _m}{\rm{ = }}\frac{1}{N}\sum\limits_{i = 1}^N {{{\left( {{e_i} - {c_m}} \right)}^2}} $ is the second order moment of error about the code word ${c_m}$.

\noindent \textit{Proof}: As $\sigma  \to \infty $, we have ${G_\sigma }\left( {{e_i} - {c_m}} \right) \approx \frac{1}{{\sqrt {2\pi } \sigma }}\left( {1 - \frac{{{{\left( {{e_i} - {c_m}} \right)}^2}}}{{2{\sigma ^2}}}} \right)$. It follows easily that

\begin{equation}\label{p4}
\begin{aligned}
\hat I_2^Q(e) &= \frac{1}{{{N^2}}}\sum\limits_{i = 1}^N {\left( {\sum\limits_{m = 1}^M {{M_m}{G_\sigma }\left( {{e_i} - {c_m}} \right)} } \right)} \\
&{\rm{       }} \approx \frac{1}{{{N^2}\sqrt {2\pi } \sigma }}\sum\limits_{i = 1}^N {\left( {\sum\limits_{m = 1}^M {{M_m}\left( {1 - \frac{{{{\left( {{e_i} - {c_m}} \right)}^2}}}{{2{\sigma ^2}}}} \right)} } \right)} \\
&{\rm{       }} = \frac{1}{{\sqrt {2\pi } \sigma }} - \frac{1}{{2{N^2}\sqrt {2\pi } {\sigma ^3}}}\sum\limits_{i = 1}^N {\left( {\sum\limits_{m = 1}^M {{M_m}{{\left( {{e_i} - {c_m}} \right)}^2}} } \right)} \\
&{\rm{       }} = \frac{1}{{\sqrt {2\pi } \sigma }} - \frac{1}{{2\sqrt {2\pi } {\sigma ^3}}}\sum\limits_{m = 1}^M {\frac{{{M_m}}}{N}\left( {\frac{1}{N}\sum\limits_{i = 1}^N {{{\left( {{e_i} - {c_m}} \right)}^2}} } \right)} \\
&{\rm{       }} = \frac{1}{{\sqrt {2\pi } \sigma }} - \frac{1}{{2\sqrt {2\pi } {\sigma ^3}}}\sum\limits_{m = 1}^M {{\alpha _m}{\mu _m}} 
\end{aligned}
\end{equation}

\textit{Remark}: By Property 4, as $\sigma  \to \infty $, the second order moments tend to dominate the QMEE cost $\hat I_2^Q(e)$. In this case, maximizing the QMEE cost is equivalent to minimizing a weighted average of the second order moments about the code words.

\noindent \textit{Property 5}: If$\forall j$, $\left| {{e_j} - Q[{e_j}]} \right| \le \varepsilon $ with $\varepsilon $ being a positive number, then  $\left| {\hat I_2^Q(e) - {{\hat I}_2}(e)} \right| \le \frac{{\varepsilon \exp ( - {1 \mathord{\left/
				{\vphantom {1 2}} \right.
				\kern-\nulldelimiterspace} 2})}}{\sigma }$.
			
\noindent \textit{Proof}: Because the Gaussian function ${G_\sigma }(.)$ is continuously differentiable over $\mathbb{R}$, according to the \textit{Mean Value Theorem}, $\forall i,j$ , there exists a point 
${\xi _{ij}} \in \left( {\min \left\{ {{e_i} - Q[{e_j}],{e_i} - {e_j}} \right\},\max \left\{ {{e_i} - Q[{e_j}],{e_i} - {e_j}} \right\}} \right)$ such that $f'\left( c \right) = \frac{{f\left( b \right) - f\left( a \right)}}{{b - a}}$.

\begin{equation}\label{p5_1}
\begin{aligned}
&{G_\sigma }\left( {{e_i} - Q[{e_j}]} \right) - {G_\sigma }\left( {{e_i} - {e_j}} \right)\\
&{\rm{     }} = {G_\sigma }\left( {{e_i} - {e_j}{\rm{ + }}\left( {{e_j} - Q[{e_j}]} \right)} \right) - {G_\sigma }\left( {{e_i} - {e_j}} \right)\\
&{\rm{      = }}\left( {{e_j} - Q[{e_j}]} \right){{G'}_\sigma }\left( {{\xi _{ij}}} \right)
\end{aligned}
\end{equation}

\noindent where ${G'_\sigma }(.)$ denotes the derivative of ${G_\sigma }(.)$ with respect to the argument. Then we have
\begin{equation}\label{p5_2}
\begin{aligned}
&\left| {{G_\sigma }\left( {{e_i} - Q[{e_j}]} \right) - {G_\sigma }\left( {{e_i} - {e_j}} \right)} \right|\\
&{\rm{     }} = \left| {{e_j} - Q[{e_j}]} \right| \times \left| {{{G'}_\sigma }\left( {{\xi _{ij}}} \right)} \right|\\
&{\rm{     }}\mathop  \le \limits^{(a)} \frac{{\varepsilon \exp ( - {1 \mathord{\left/
				{\vphantom {1 2}} \right.
				\kern-\nulldelimiterspace} 2})}}{\sigma }
\end{aligned}
\end{equation}

\noindent where (a) comes from $\left| {{e_j} - Q[{e_j}]} \right| \le \varepsilon $ and $\left| {{{G'}_\sigma }\left( x \right)} \right| \le \frac{{\exp ( - {1 \mathord{\left/
				{\vphantom {1 2}} \right.
				\kern-\nulldelimiterspace} 2})}}{\sigma }$ for any $x \in \mathbb{R}$. It follows that 

\begin{equation}\label{p5_3}
\begin{aligned}
\left| {\hat I_2^Q(e) \!-\! {{\hat I}_2}(e)} \right| &\!=\! \left| {\frac{1}{{{N^2}}}\sum\limits_{i = 1}^N {\sum\limits_{j = 1}^N {\left( {{G_\sigma }\left( {{e_i} \!-\! Q[{e_j}]} \right) \!-\! {G_\sigma }\left( {{e_i} \!-\! {e_j}} \right)} \right)} } } \right|\\
&{\rm{                     }} \le \frac{1}{{{N^2}}}\sum\limits_{i = 1}^N {\sum\limits_{j = 1}^N {\left| {{G_\sigma }\left( {{e_i} \!-\! Q[{e_j}]} \right) \!-\! {G_\sigma }\left( {{e_i} \!-\! {e_j}} \right)} \right|} } \\
&{\rm{                     }} \le \frac{1}{{{N^2}}}\sum\limits_{i = 1}^N {\sum\limits_{j = 1}^N {\frac{{\varepsilon \exp ( - {1 \mathord{\left/
						{\vphantom {1 2}} \right.
						\kern-\nulldelimiterspace} 2})}}{\sigma }} } \\
&{\rm{                     }} = \frac{{\varepsilon \exp ( - {1 \mathord{\left/
				{\vphantom {1 2}} \right.
				\kern-\nulldelimiterspace} 2})}}{\sigma }
\end{aligned}
\end{equation}

\textit{Remark}: From Property 5, when $\varepsilon $ is very small or $\sigma$ is very large, the difference between the values of $\hat I_2^Q(e)$ and ${\hat I_2}(e)$ will be very small.

\noindent \textit{Property 6}: For a linear regression model $f(x) = {\omega ^T}x$, with $\omega  \in {\mathbb{R}^d}$ being the weight  vector  to be estimated, the optimal solution under QMEE criterion satisfies  

\begin{equation}\label{p6_1}
\omega  = R_{QMEE}^{ - 1}{P_{QMEE}}
\end{equation}

\noindent where ${R_{QMEE}} = \sum\limits_{i = 1}^N {\sum\limits_{m = 1}^M {{M_m}{G_\sigma }\left( {{e_i} - {c_m}} \right){x_i}{x_i}^T} } $ and ${P_{QMEE}} = \sum\limits_{i = 1}^N {\sum\limits_{m = 1}^M {{M_m}{G_\sigma }\left( {{e_i} - {c_m}} \right)({y_i} - {c_m}){x_i}} } $.

\noindent \textit{Proof}: The derivative of the QMEE cost $\hat I_2^Q(e)$ with respect to $\omega$ is

\begin{equation}\label{p6_2}
\begin{aligned}
\frac{\partial }{{\partial \omega }}\hat I_2^Q(e) =& \frac{1}{{{N^2}}}\sum\limits_{i = 1}^N {\left( {\sum\limits_{m = 1}^M \!\!{{M_m}\frac{\partial }{{\partial \omega }}{G_\sigma }\left( {{e_i} - {c_m}} \right)} } \right)} \\
{\rm{               }} =& \frac{1}{{{N^2}{\sigma ^2}}}\sum\limits_{i = 1}^N {\!\!\left( {\sum\limits_{m = 1}^M {{M_m}{G_\sigma }\left( {{e_i} \!-\! {c_m}} \right)({y_i} \!-\! {\omega ^T}{x_i} \!-\! {c_m}){x_i}} } \!\!\right)} \\
=& \frac{1}{{{N^2}{\sigma ^2}}} {\sum\limits_{i = 1}^N {\sum\limits_{m = 1}^M {{M_m}{G_\sigma }\left( {{e_i} - {c_m}} \right)({y_i} - {c_m}){x_i}} }}  \\
&-  \frac{1}{{{N^2}{\sigma ^2}}} \left( {\sum\limits_{i = 1}^N {\sum\limits_{m = 1}^M {{M_m}{G_\sigma }\left( {{e_i} \!-\! {c_m}} \right){x_i}{x_i}^T} } } \right)\omega  \\
{\rm{               }} =& \frac{1}{{{N^2}{\sigma ^2}}}\left\{ {{P_{QMEE}} - {R_{QMEE}}\omega } \right\}
\end{aligned}
\end{equation}

\noindent Setting $\frac{\partial }{{\partial \omega }}\hat I_2^Q(e) = 0$, we get $\omega  = R_{QMEE}^{ - 1}{P_{QMEE}}$. It completes the proof.

\textit{Remark}: It is worth noting that the solution $\omega  = R_{QMEE}^{ - 1}{P_{QMEE}}$ is not a closed-form solution as the matrix 
$ {R_{QMEE}} $ and the vector $ {P_{QMEE}} $ on the right side of the equation depend on the weight vector $\omega$ through the error samples (i.e. $ {e_i} = {y_i} - {\omega ^T}{x_i} $ ). Actually, the equation $\omega  = R_{QMEE}^{ - 1}{P_{QMEE}}$ is a fixed-point equation.

A key problem in QMEE is how to design a simple and efficient quantizer $Q[.]$, including how to build the codebook and how to assign the code words to the data. In this work, we will use a method proposed in our recent papers, to quantize the error samples. In \cite{chen2012quantized,chen2013quantized}, we proposed a simple online vector quantization (VQ) to curb the network growth in kernel adaptive filters, such as kernel least mean square (KLMS) and kernel recursive least squares (KRLS). The main advantage of this quantization method lies in its simplicity and online feature. The pseudocode of this online VQ algorithm is presented in \textit{Algorithm 1}.

\begin{algorithm}
	\renewcommand{\algorithmicrequire}{\textbf{Input:}}
	\renewcommand{\algorithmicensure}{\textbf{Output:}}
	\caption{}
	\label{alg:1}
	\begin{algorithmic}[1]
		\REQUIRE error samples $ \{ {e_i}\} _{i = 1}^N $
		\ENSURE quantized errors $ \{ Q[{e_i}]\} _{i = 1}^N $
		\STATE Parameters setting: quantization threshold $ \varepsilon $  
		\STATE Initialization: Set ${C_1} = \left\{ {{e_1}} \right\}$, where ${C_i}$ denotes the codebook at the iteration $i$ 
		\FOR{$ i = 2,...,N $}
		\STATE Compute the distance between $ {e_i} $ and $ {C_{i-1}} $:\\
		$dis\left( {{e_i},{C_{i - 1}}} \right) = \left| {{e_i} - {C_{i - 1}}({j^*})} \right|$\\
		where ${j^*} = \mathop {{\mathop{\rm argmin}\nolimits} }\limits_{1 \le j \le \left| {{C_{i - 1}}} \right|} \left| {{e_i} - {C_{i - 1}}(j)} \right|$, ${C_{i - 1}}(j)$ denotes the $j$th element of ${C_{i - 1}}$, and $\left| {{C_{i - 1}}} \right|$ stands for the cardinality of ${C_{i - 1}}$.
		\IF {$dis\left( {{e_i},{C_{i - 1}}} \right) \le \varepsilon $}
		\STATE Keep the codebook unchanged: ${C_i} = {C_{i - 1}}$ and quantize ${e_i}$ to the closest code word $Q[{e_i}] = {C_{i - 1}}({j^*})$;
		\ELSE
		\STATE Update the codebook: ${C_i} = \left\{ {{C_{i - 1}},{e_i}} \right\}$ and quantize ${e_i}$ to itself: $Q[{e_i}] = {e_i}$;
		\ENDIF
		\ENDFOR
	\end{algorithmic}  
\end{algorithm}

\textit{Remark}: The online VQ method in Algorithm 1 creates the codebook sequentially from the samples, which is computationally very simple, with computational complexity that is linear in the number of samples.

\section{ILLUSTRATIVE EXAMPLES}
In the following, we present some illustrative examples to demonstrate the desirable performance of the proposed QMEE criterion.
\subsection{Linear Regression}
In the first example, we use the QMEE criterion to perform the linear regression. According to Property 6, the optimal solution of the linear regression model $f(x) = {\omega ^T}x$ can easily be solved by the following fixed-point iteration:

\begin{equation}\label{lr}
{\omega _k} = {\left[ {{R_{QMEE}}({\omega _{k - 1}})} \right]^{ - 1}}{P_{QMEE}}({\omega _{k - 1}})
\end{equation}

\noindent in which the matrix ${R_{QMEE}}({\omega _{k - 1}})$ and vector ${P_{QMEE}}({\omega _{k - 1}})$ are

\begin{equation}\label{rp}
\left\{ \begin{array}{l}
{R_{QMEE}}({\omega _{k - 1}}) = \sum\limits_{m = 1}^M {\textbf{X}{\bf{\Lambda} _m}{\textbf{X}^T}} \\
{P_{QMEE}}({\omega _{k - 1}}) = \sum\limits_{m = 1}^M {\textbf{X}{\bf{\Lambda} _m}{\textbf{Y}_m}} 
\end{array} \right.
\end{equation}

\noindent where $\textbf{X} = \left[ {{x_1},{x_2}, \cdots ,{x_N}} \right] \in {\mathbb{R}^{d \times N}}$, ${\textbf{Y}_m} = {\left[ {{y_1} - {c_m},{y_2} - {c_m}, \cdots ,{y_N} - {c_m}} \right]^T} \in {\mathbb{R}^N}$, and ${\bf{\Lambda} _m}$ is a $N\times N$ diagonal matrix with diagonal elements ${\bf{\Lambda} _m}(ii) = {M_m}{G_\sigma }\left( {{e_i} - {c_m}} \right)$ , with  . The detailed procedure of the linear regression under QMEE is summarized in \textit{Algorithm 2}.

\begin{algorithm}
	\renewcommand{\algorithmicrequire}{\textbf{Input:}}
	\renewcommand{\algorithmicensure}{\textbf{Output:}}
	\caption{}
	\label{alg:1}
	\begin{algorithmic}[1]
		\REQUIRE samples $\{ {x_i},{y_i}\} _{i = 1}^N$
		\ENSURE weight vector $\omega$ 
		\STATE Parameters setting: iteration number $K$, kernel width $\sigma$, quantization threshold $\varepsilon $   
		\STATE Initialization: Set $\omega_0=\textbf{0}$
		\FOR{$ k = 2,...,K $}
		\STATE Compute the error samples based on $\omega_{k - 1}$: ${e_i} = {y_i} - \omega _{k - 1}^T{x_i}$, $i = 1,2, \cdots ,N$ ;
		\STATE Create the quantization codebook $C$ and quantize the $N$ error samples by \textit{Algorithm 1};
		\STATE Compute the matrix ${R_{QMEE}}({\omega _{k - 1}})$ and the vector ${P_{QMEE}}({\omega _{k - 1}})$ by (\ref{rp});
		\STATE Update the weight vector by (\ref{lr});
		\ENDFOR
	\end{algorithmic}  
\end{algorithm}

We now consider a simple scenario where the data samples are generated by a two-dimensional linear system ${y_i} = {\omega ^*}^T{x_i} + {v_i}$, where ${\omega ^*} = {\left[ {2,1} \right]^T}$, and ${v_i}$ is an additive noise. The input vectors $\left\{ {{x_i}} \right\}$ are assumed to be uniformly distributed over $\left[ { - 2,2} \right] \times \left[ { - 2,2} \right]$. In addition, the noise ${v_i}$ is assumed to be generated by ${v_i} = \left( {1 - {a_i}} \right){A_i} + {a_i}{B_i}$, where ${a_i}$ is a binary process with probability mass $\Pr \left\{ {{a_i} = 1} \right\} = c$, $\Pr \left\{ {{a_i} = 0} \right\} = 1 - c$, with $0 \le c \le 1$ being an occurrence probability. The processes ${A_i}$ and ${B_i}$ represent the background noises and the outliers respectively, which are mutually independent and both independent of ${a_i}$. In the simulations below, $c$ is set at 0.1 and ${B_i}$ is assumed to be a white Gaussian process with zero-mean and variance 10000. For the distribution of ${A_i}$, we consider four cases: 1) symmetric Gaussian mixture density: $0.5\mathcal{N}\left( {3,1} \right) + 0.5\mathcal{N}\left( { - 3,1} \right)$, where $\mathcal{N}\left( {\mu ,{\sigma ^2}} \right)$ denotes the Gaussian density with mean $\mu$ and variance $\sigma ^2$; 2) asymmetric Gaussian mixture density: $\frac{2}{3}\mathcal{N}\left( { - 5,1} \right) + \frac{1}{3}\mathcal{N}\left( {2,1} \right)$ ; 3) binary distribution with probability mass $\Pr \left\{ {x =  - 2} \right\} = \Pr \left\{ {x = 2} \right\} = 0.5$; 4) Gaussian distribution with zero-mean and unit variance. The root mean squared error (RMSE) is employed to measure the performance, computed by

\begin{equation}\label{rmse}
RMSE = \sqrt {\frac{1}{2}{{\left\| {{\omega _k} - {\omega ^*}} \right\|}^2}} 
\end{equation}

\noindent where $\omega_k$ and ${\omega ^*}$ denote the estimated and the target weight vectors respectively.

We compare the performance of four learning criteria, namely MSE, MCC \cite{liu2007correntropy,he2011robust,chen2012maximum,chen2015convergence,zhao2011kernel}, MEE and QMEE. For the MSE criterion, there is a closed-form solution, so no iteration is needed. For other three criteria, a fixed-point iteration is used to solve the model (see \cite{chen2015convergence,zhang2015convergence} for the details of the fixed-point algorithms under MCC and MEE). The parameter settings of MCC, MEE and QMEE are given in Table I. The simulations are carried out with MATLAB 2014a running in i5-4590, 3.30 GHZ CPU. The “mean ±deviation” results of the RMSE and the training time over 100 Monte Carlo runs are presented in Table II. In the simulations, the sample number is $N = 200$ and the iteration number is $K=100$. From Table II, we observe: i) the MCC, MEE and QMEE can significantly outperform the traditional MSE criterion although they have no closed-form solution; ii) the MEE and QMEE can achieve much better performance than the MCC criterion, except the case of Gaussian background noise, in which they achieve almost the same performance; iii) the QMEE can achieve almost the same (or even better) performance as the original MEE criterion, but with much less computational cost. Fig. 1 shows the average training time of QMEE and MEE with increasing number of samples.

Further, we show in Fig. 2 the contour plots of the performance surfaces (i.e. the cost surfaces over the parameter space), where the background noise distribution is assumed to be symmetric Gaussian mixture. In Fig. 2, the target weight vector and the optimal solutions of the performance surfaces are denoted by the red crosses and blue circles, respectively. As one can see, the optimal solutions under MEE and QMEE are almost identical to the target value, while the solutions under MSE and MCC (especially the MSE solution) are apart from the target.

\begin{table}[htbp]\small
	\renewcommand\arraystretch{1.5}
	\setlength{\abovecaptionskip}{0pt}
	\setlength{\belowcaptionskip}{5pt}
	\centering
	\caption{Parameter settings of three criteria}
	\begin{tabular}{cccccc}
		\toprule
		&$\quad$&MCC&MEE&\multicolumn{2}{c}{QMEE}\\
		\hline
		&$\quad$&$\sigma$&$\sigma$&$\sigma$&$\epsilon$\\
		\hline
		&Case 1)&10&1.1&1.5&0.3\\
		&Case 2)&15&1.1&1.5&0.3\\
		&Case 3)&8&0.7&1.0&0.3\\
		&Case 4)&2.8&0.6&4.0&0.1\\
		\bottomrule
	\end{tabular}
\end{table}

\begin{table*}[htbp]\footnotesize
	\renewcommand\arraystretch{1.5}
	\setlength{\abovecaptionskip}{0pt}
	\setlength{\belowcaptionskip}{5pt}
	\centering
	\caption{RMSE and training time of different criteria}
	\begin{tabular}{cccccccc}
		\toprule
		&$\quad$&MSE&MCC&MEE&QMEE\\
		\hline
		\multirow{2}*{Case 1)} &RMSE&$1.1649\pm0.6587$&$ 0.1493\pm0.0756 $&$ 0.0468\pm0.0205 $&$ 0.0473\pm0.0203 $\\
		{}&Training Time (sec)&$N/A$&$3.0000 \!\!\times\!\! {{10}^{ - 4}} \pm 2.6000 \!\!\times\!\! {{10}^{ - 4}}$&$0.2963 \pm 3.5300 \!\!\times\!\! {{10}^{ - 3}}$&$9.1410 \!\!\times\!\! {{10}^{ - 3}} \pm 6.1800 \!\!\times\!\! {{10}^{ - 4}}$\\
		\hline
		\multirow{2}*{Case 2)} &RMSE&$1.2951 \pm 0.6701$&$ 0.1987 \pm 0.1111 $&$ 0.0455 \pm 0.0226 $&$ 0.0460 \pm 0.0227 $\\
		{}&Training Time (sec)&$N/A$&$3.3900 \!\!\times\!\! {{10}^{ - 4}} \pm 2.6000 \!\!\times\!\! {{10}^{ - 4}}$&$0.3013 \pm 8.4110 \!\!\times\!\! {{10}^{ - 3}}$&$9.0140 \!\!\times\!\! {{10}^{ - 3}} \pm 8.6000 \!\!\times\!\! {{10}^{ - 4}}$\\
		\hline
		\multirow{2}*{Case 3)} &RMSE&$1.0939 \pm 0.6407$&$ 0.0928 \pm 0.0480 $&$ 7.7500\times10^{-4} \pm 0.0010 $&$ 7.8940\times10^{-4} \pm 0.0010 $\\
		{}&Training Time (sec)&$N/A$&$3.6700 \!\!\times\!\! {{10}^{ - 4}} \pm 2.6500 \!\!\times\!\! {{10}^{ - 4}}$&$0.2932 \pm 3.4600 \!\!\times\!\! {{10}^{ - 3}}$&$7.3230 \!\!\times\!\! {{10}^{ - 3}} \pm 5.2700 \!\!\times\!\! {{10}^{ - 4}}$\\
		\hline
		\multirow{2}*{Case 4)} &RMSE&$1.2031 \pm 0.6531$&$ 0.0422 \pm 0.0224 $&$ 0.0452\pm 0.0262 $&$ 0.0422 \pm 0.0231 $\\
		{}&Training Time (sec)&$N/A$&$3.5300 \!\!\times\!\! {{10}^{ - 4}} \pm 2.6100 \!\!\times\!\! {{10}^{ - 4}}$&$0.2999 \pm 2.4750 \!\!\times\!\! {{10}^{ - 3}}$&$7.9500 \!\!\times\!\! {{10}^{ - 3}} \pm 6.4300 \!\!\times\!\! {{10}^{ - 4}}$\\
		\bottomrule
	\end{tabular}
\end{table*}

\begin{figure}[htbp]
	\setlength{\abovecaptionskip}{0pt}
	\setlength{\belowcaptionskip}{0pt}
	\centering
	\includegraphics[width=3.0in,height=2.4in]{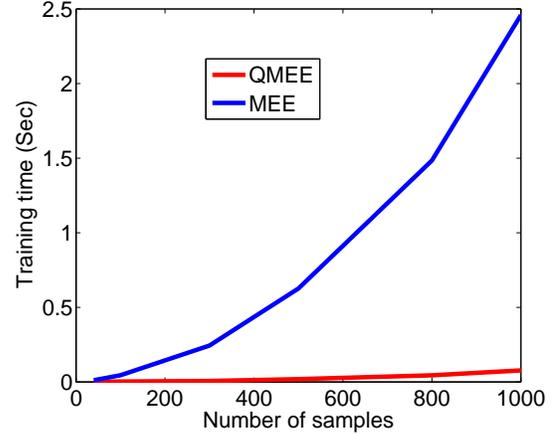}
	\caption{Training time versus the number of samples}
	\label{fig1}
\end{figure}

\begin{figure*}[htbp]
	\setlength{\abovecaptionskip}{0pt}
	\setlength{\belowcaptionskip}{0pt}
	%\makeatletter
	%\def\@captype{figure}
	%\makeatother
	%\begin{minipage}[t]{0.5\linewidth}
	\centering
	\subfigure[]{
		\includegraphics[width=3.0in,height=2.4in]{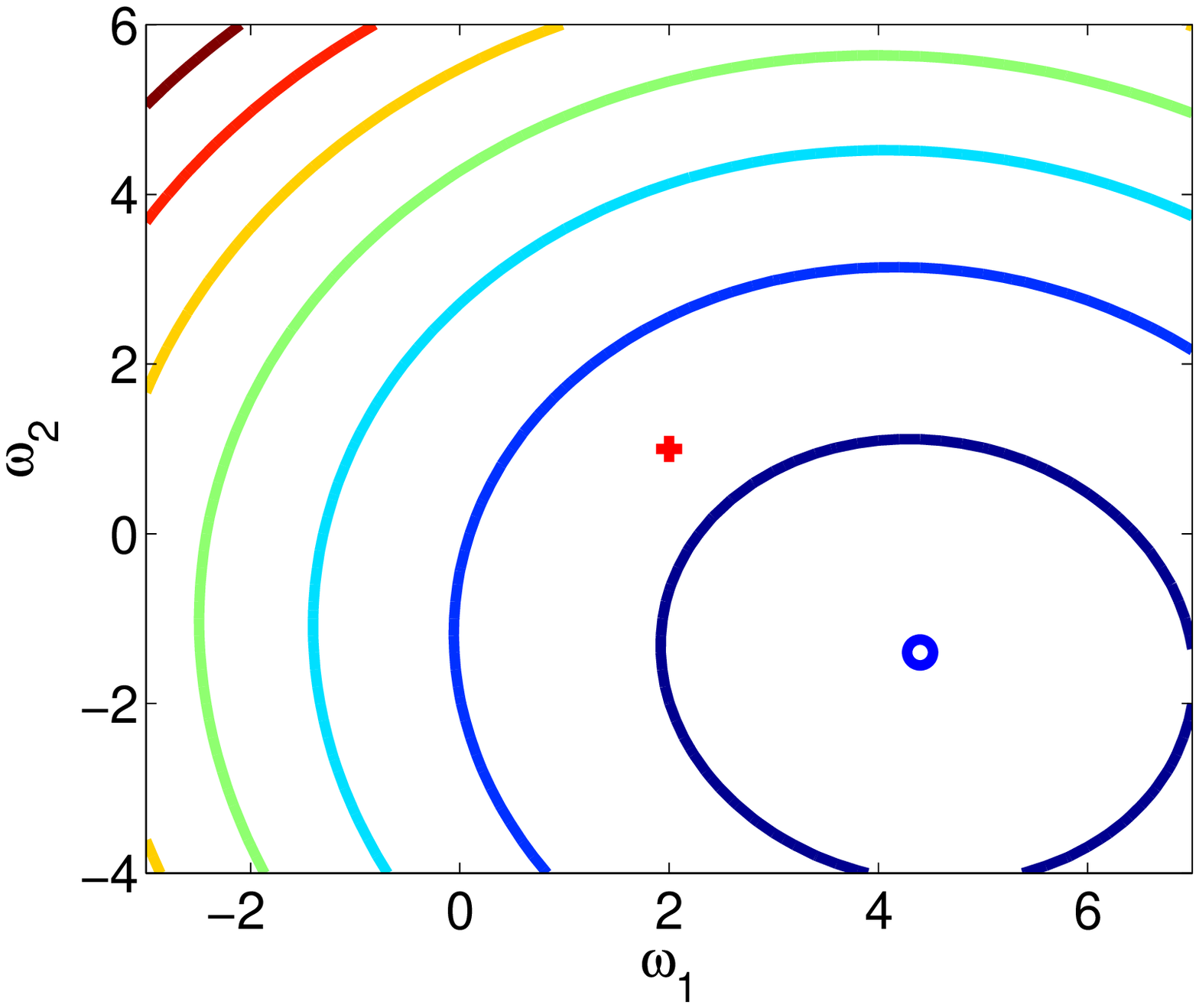}}
	\subfigure[]{
		\includegraphics[width=3.0in,height=2.4in]{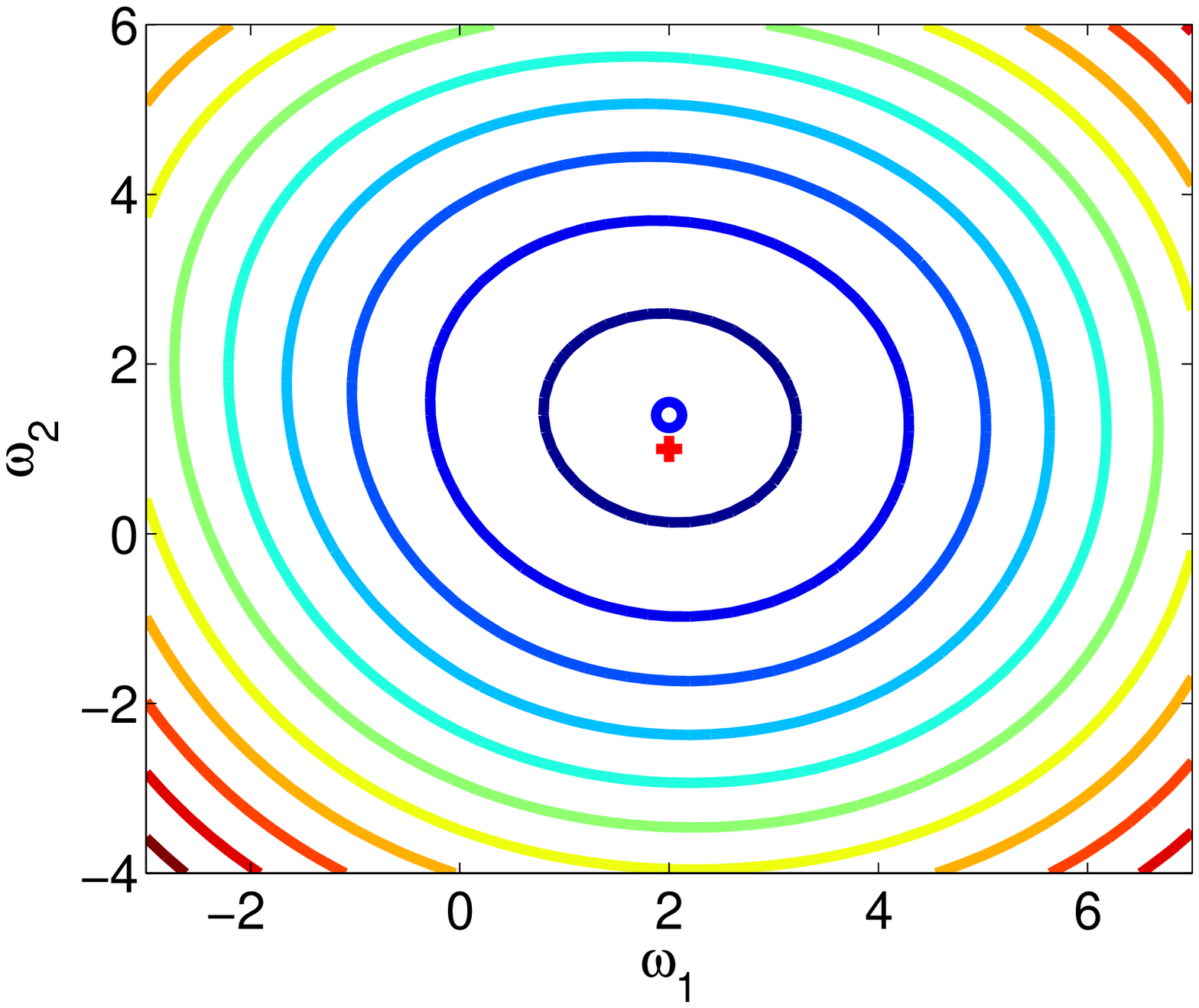}}
	\subfigure[]{
		\includegraphics[width=3.0in,height=2.4in]{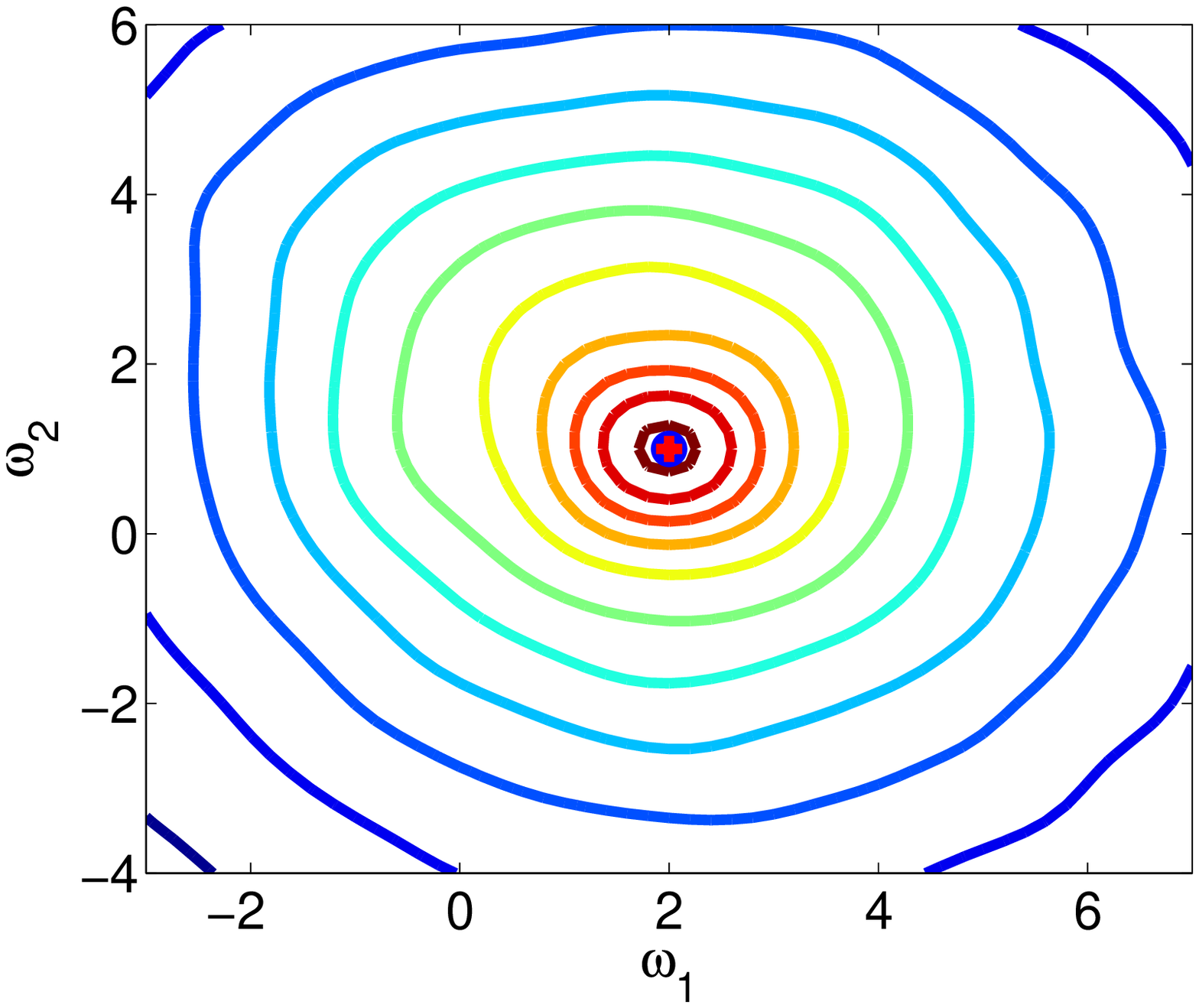}}
	\subfigure[]{
		\includegraphics[width=3.0in,height=2.4in]{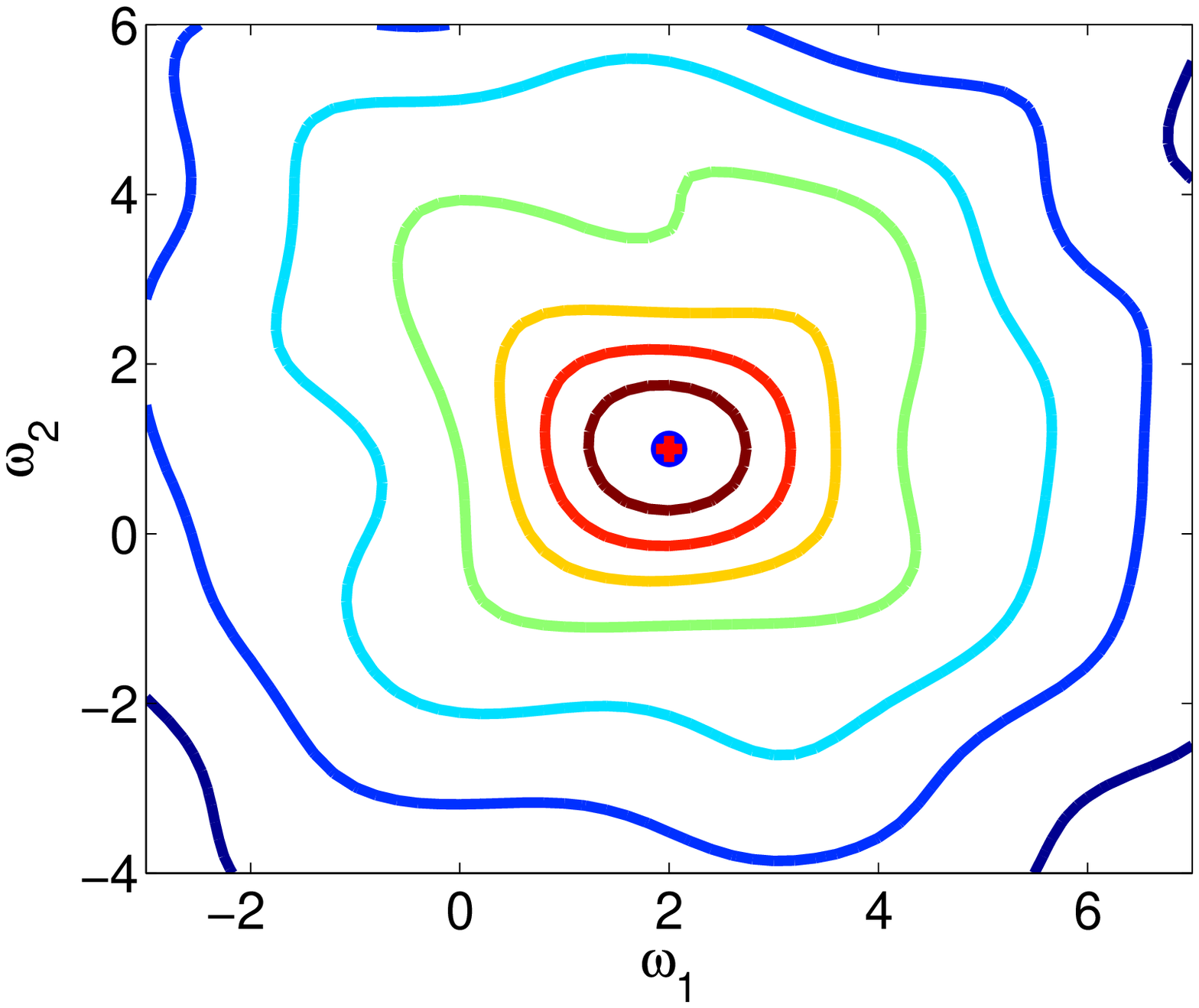}}
	\caption{Contour plots of the performance surfaces (a) MSE; (b) MCC; (c) MEE; (d) QMEE}
	\label{fig2}
\end{figure*}

\subsection{Extreme Learning Machines}
The second example is about the training of \textit{Extreme Learning Machine} (ELM) \cite{huang2006extreme,huang2012extreme,huang2017efficient,xing2013training,yang2015data}, a single-hidden-layer feedforward neural network (SLFN) with random hidden nodes.

Given $N$ distinct training samples $\left\{ {{\textbf{x}_i},{t_i}} \right\}_{i = 1}^N$, with ${\textbf{x}_i} = {\left[ {{x_{i1}},{x_{i2}}, \ldots ,{x_{id}}} \right]^T} \in {\mathbb{R}^d}$ being the input vector and $t_i \in \mathbb{R}$ the target response, the output of a standard SLFN with $L$ hidden nodes is

\begin{equation}\label{slfn}
{y_i} = \sum\limits_{j = 1}^L {{\beta _j}f\left( {{\textbf{w}_j}{\textbf{x}_i} + {b_j}} \right)} 
\end{equation}

\noindent where $f(.)$ is an activation function, ${\textbf{w}_j} = [{w_{j1}},{w_{j2}},...,{w_{jd}}] \in {\mathbb{R}^d}$ and $b_j \in \mathbb{R}$ ($j = 1,2,...,L$ ) are the randomly generated parameters of the $L$ hidden nodes, and $\bm{\beta}  = {({\beta _1},...,{\beta _L})^T} \in {\mathbb{R}^L}$ represents the output weight vector. Since the hidden parameters are determined randomly, we only need to solve the output weight vector $\bm{\beta}$. To this end, we express (\ref{slfn}) in a vector form as

\begin{equation}\label{slfn}
\textbf{Y}=\textbf{H}\bm{\beta}
\end{equation}

\noindent where ${\bf{Y}} = {({y_1},...,{y_N})^T}$, and

\begin{equation}\label{H}
{\bf{H}} = \left( {\begin{array}{*{20}{c}}
	{{\textbf{h}_1}}\\
	\vdots \\
	{{\textbf{h}_N}}
	\end{array}} \right) = \left( \begin{array}{l}
f({\textbf{w}_1}{\textbf{x}_1} + {b_1}),\;\;\;...\;\;\\
\;\;\;\;\;\;\;\;\; \vdots \;\;\;\;\;\;\;\;\;\;\;\;\; \ddots \\
f({\textbf{w}_1}{\textbf{x}_N} + {b_1}),\;\;\;...\;
\end{array} \right.\left. \begin{array}{l}
f({\textbf{w}_L}{\textbf{x}_1} + {b_L})\\
\;\;\;\;\;\;\;\;\;\; \vdots \\
f({\textbf{w}_L}{\textbf{x}_N} + {b_L})
\end{array} \right)
\end{equation}

\noindent Usually, the output weight vector $\bm{\beta}$ can be solved by minimizing the following squared (MSE based) and regularized loss function:

\begin{equation}\label{relm_mse}
{J_{MSE}}\left( \bm{\beta}  \right) = \sum\limits_{i = 1}^N {e_i^2}  + \lambda \left\| \beta  \right\|_2^2 = \left\| {\textbf{H}\bm{\beta}  - \textbf{T}} \right\|_2^2 + \lambda \left\| \bm{\beta}  \right\|_2^2
\end{equation}

\noindent where ${e_i} = {t_i} - {y_i} = {t_i} - {\textbf{h}_i}\bm{\beta}$ is the $i$th error between the target response and actual output, $\lambda  \ge 0$ represents the regularization factor, and $\textbf{T} = {\left( {{t_1},...,{t_N}} \right)^T}$ . Applying the pseudo inversion operation, one can obtain a unique solution under the loss function (\ref{relm_mse}), that is 

\begin{equation}\label{relm_solu}
\bm{\beta}  = {\left[ {{\textbf{H}^T}\textbf{H} + \lambda \textbf{I}} \right]^{ - 1}}{\textbf{H}^T}\textbf{T}
\end{equation}

Here, we propose the following QMEE based loss function:

\begin{equation}\label{elm_qmee}
\begin{aligned}
&{J_{QMEE}}\left( \beta  \right) =  - {{\hat I}_2}(e) + \lambda \left\| \beta  \right\|_2^2\\
&=  - \frac{1}{{{N^2}}}\sum\limits_{i = 1}^N {\left( {\sum\limits_{m = 1}^M {{M_m}{G_\sigma }\left( {{e_i} - {c_m}} \right)} } \right)}  + \lambda \left\| \beta  \right\|_2^2\\
&=  - \frac{1}{{{N^2}}}\sum\limits_{i = 1}^N {\left( {\sum\limits_{m = 1}^M {{M_m}\exp \left( { - \frac{{{{\left( {{e_i} - {c_m}} \right)}^2}}}{{2{\sigma ^2}}}} \right)} } \right)}  + \lambda \left\| \beta  \right\|_2^2
\end{aligned}
\end{equation}

\noindent Setting $\frac{{\partial {J_{QMEE}}(\beta )}}{{\partial \beta }} = 0$, one can obtain
\begin{equation}\label{elm_qmee_solu}
\bm{\beta}  = {[A + \lambda '{\bf{I}}]^{ - 1}}B
\end{equation}

\noindent where $A = \sum\limits_{m = 1}^M {{\textbf{H}^T}{\bf{\Lambda} _m}\textbf{H}} $ , $ B = \sum\limits_{m = 1}^M {{\textbf{H}^T}{\bf{\Lambda} _m}{\textbf{T}_m}}$ ,$\lambda ' = 2\lambda {N^2}{\sigma ^2}$ , ${\textbf{T}_m} = {\left[ {{t_1} - {c_m}, \cdots ,{t_N} - {c_m}} \right]^T}$ , and $ \bf{\Lambda}_m $ is a $N \times N$ diagonal matrix with diagonal elements ${\bf{\Lambda} _m}(ii) = {M_m}{G_\sigma }\left( {{e_i} - {c_m}} \right)$ .

Similar to the linear regression case, the equation (\ref{elm_qmee_solu}) is a fixed-point equation since the matrix $\bf{\Lambda}_m$  depends on the weight vector $\bm{\beta}$ through ${e_i} = {t_i} - {\textbf{h}_i}\bm{\beta} $. Thus, one can solve $\bm{\beta}$ by using the following fixed-point iteration:

\begin{equation}\label{26_fix_point}
{\bm{\beta} _k} = {[A({\bm{\beta} _{k - 1}}) + \lambda '{\bf{I}}]^{ - 1}}B({\bm{\beta} _{k - 1}})
\end{equation}

\noindent where $A({\bm{\beta} _{k - 1}})$ and $B({\bm{\beta} _{k - 1}})$ denote, respectively, the matrix $A$ and vector $B$ evaluated at ${\bm{\beta} _{k - 1}}$. The learning procedure of the ELM under QMEE is described in \textit{Algorithm 3}. This algorithm is called the ELM-QMEE in this paper.

\begin{algorithm}
	\renewcommand{\algorithmicrequire}{\textbf{Input:}}
	\renewcommand{\algorithmicensure}{\textbf{Output:}}
	\caption{ELM-QMEE}
	\label{alg:1}
	\begin{algorithmic}[1]
		\REQUIRE samples $\{ {x_i},{y_i}\} _{i = 1}^N$
		\ENSURE weight vector $\bm{\beta}$ 
		\STATE Parameters setting: number of hidden nodes $L$, regularization parameter $\lambda '$, iteration number $K$, kernel width $\sigma$, quantization threshold $\varepsilon $    
		\STATE Initialization: set $\bm{\beta}_0=\textbf{0}$ and randomly initialize the hidden parameters ${\textbf{w}_j}$ and $b_j$ ($j = 1,...,L$ )
		\FOR{$ k = 2,...,K $}
		\STATE Compute the error samples based on $\bm{\beta}_{k - 1}$: ${e_i} = {t_i} - {\textbf{h}_i}{\bm{\beta} _{k - 1}}$, $i = 1,2, \cdots ,N$ ;
		\STATE Create the quantization codebook $C$ and quantize the $N$ error samples by \textit{Algorithm 1};
		\STATE Compute the matrix $A({\bm{\beta} _{k - 1}})$ and the vector $B({\bm{\beta} _{k - 1}})$ by (\ref{rp});
		\STATE Update the weight vector $\bm{\beta}$ by (\ref{26_fix_point});
		\ENDFOR
	\end{algorithmic}  
\end{algorithm}

In the following, we consider the regression problem with five benchmark datasets from the UCI machine learning repository \cite{frank2010uci}.The details of the datasets are shown in Table III. For each dataset, the training and testing samples are randomly selected form the dataset. Particularly, the data are normalized to the range [0, 1]. Five algorithms are compared here, including ELM \cite{huang2006extreme}, RELM \cite{huang2012extreme}, ELM-RCC \cite{xing2013training}, ELM-MEE and ELM-QMEE. The ELM-MEE can be viewed as the ELM-QMEE with $\varepsilon  = 0$. The parameter settings of the five ELM algorithms are presented in Table IV, which are experimentally chosen by fivefold cross-validation. 

The RMSE is used as the performance measure for regression. The ``mean $\pm$ standard deviation'' results of Testing RMSE and the Training time over 100 runs are shown in Table V and VI. In addition, since the MEE and QMEE criteria are shift-invariant, the RMSE of MEE and QMEE are calculated by adding a bias value to the testing errors. This bias value was adjusted so as to yield zero-mean error over the training set. As one can see, in all the cases the proposed ELM-QMEE can outperform other algorithms except the ELM-MEE, and the results of ELM-QMEE is very close to those of ELM-MEE. Besides, compared with ELM-MEE, the computational complexity of ELM-QMEE is much smaller.

\begin{table}[]\small
	\renewcommand\arraystretch{1.5}
	\setlength{\abovecaptionskip}{0pt}
	\setlength{\belowcaptionskip}{0pt}
	\centering
	\caption{Specification of the datasets}
	\begin{tabular}{cccc}
		\toprule
		\multirow{2}*{Datasets} & \multirow{2}*{Features} & \multicolumn{2}{c}{Observations}\\
		\cline{3-4}
		&{}&Training&Testing\\
		\midrule
		Servo	&5	&83	&83\\
		Yacht	&6	&154	&154\\
		Computer Hardware  &8  &105    &104\\
		Price   &16   &80   &79\\
		Machine-CPU   &6   &105   &104\\
		\bottomrule
	\end{tabular}
\end{table}

\begin{table*}[]\small
	\renewcommand\arraystretch{1.5}
	\setlength{\abovecaptionskip}{0pt}
	\setlength{\belowcaptionskip}{5pt}
	\centering
	\caption{Parameter settings of five ELM algorithms}
	\begin{tabular}{cccccccccccccc}
		\toprule
		\multirow{2}*{Datasets} & ELM & \multicolumn{2}{c}{RELM}& \multicolumn{3}{c}{ELM-RCC}& \multicolumn{3}{c}{ELM-MEE}&\multicolumn{4}{c}{ELM-QMEE}\\
		\cline{2-14}
		&L&L&$\lambda$&L&$\sigma$&$\lambda$&L&$\sigma$&$\lambda$&L&$\sigma$&$\lambda '$&$\gamma $\\
		\hline
		Servo	&25&90&$1 \times 10^{-5}$&65&0.8&$1 \times 10^{-4}$&55&0.1&$8 \times 10^{-7}$&75&0.2&$4 \times 10^{-4}$&0.05\\
		Yacht	&90&187&$2.5 \times 10^{-5}$&195&0.4&$1 \times 10^{-7}$&225&0.1&$1 \times 10^{-7}$&210&0.2&$5 \times 10^{-7}$&0.6\\
		Computer Hardware	&20&35&$9 \times 10^{-6}$&40&0.1&$8 \times 10^{-6}$&95&0.2&$5 \times 10^{-6}$&90&0.1&$6 \times 10^{-5}$&0.009\\
		Price	&20&20&$4 \times 10^{-6}$&15&0.3&$4 \times 10^{-5}$&15&0.3&$7 \times 10^{-6}$&15&0.3&$8 \times 10^{-6}$&0.02\\
		Machine-CPU	&10&30&$7 \times 10^{-5}$&20&0.2&$7 \times 10^{-5}$&25&0.3&$5 \times 10^{-6}$&25&0.4&$0.06$&0.08\\
		\bottomrule
	\end{tabular}
\end{table*}

\begin{table*}[]\small
	\renewcommand\arraystretch{1.5}
	\setlength{\abovecaptionskip}{0pt}
	\setlength{\belowcaptionskip}{5pt}
	\centering
	\caption{Testing RMSE of five ELM algorithms}
	\begin{tabular}{cccccc}
		\toprule
		\multicolumn{1}{c}{\multirow{1}*{Datasets}} & \multicolumn{1}{c}{ELM} & \multicolumn{1}{c}{RELM}& \multicolumn{1}{c}{ELM-RCC}& \multicolumn{1}{c}{ELM-MEE}& \multicolumn{1}{c}{ELM-QMEE}\\
		
		\hline
		\multicolumn{1}{c}{\multirow{1}*{Servo}}
		&\multicolumn{1}{c}{0.1199$\pm$0.0200}
		&\multicolumn{1}{c}{0.1046$\pm$0.0178}
		&\multicolumn{1}{c}{0.1029$\pm$0.0158}
		&\multicolumn{1}{c}{\textbf{0.1014}$\pm$\textbf{0.0194}}
		&\multicolumn{1}{c}{0.1014$\pm$0.0196}\\
	
		\multicolumn{1}{c}{\multirow{1}*{Yacht}}
		&\multicolumn{1}{c}{0.0596$\pm$0.0171}
		&\multicolumn{1}{c}{0.0490$\pm$0.0058}
		&\multicolumn{1}{c}{0.1029$\pm$0.0158}
		&\multicolumn{1}{c}{0.0327$\pm$0.0080}
		&\multicolumn{1}{c}{\textbf{0.0223}$\pm$\textbf{0.0108}}\\
	
		\multicolumn{1}{c}{\multirow{1}*{Computer Hardware}}
		&\multicolumn{1}{c}{0.0262$\pm$0.0198}
		&\multicolumn{1}{c}{0.0170$\pm$0.0110}
		&\multicolumn{1}{c}{0.0162$\pm$0.0125}
		&\multicolumn{1}{c}{\textbf{0.0140}$\pm$\textbf{0.0081}}
		&\multicolumn{1}{c}{0.0147$\pm$0.0114}\\
	
		\multicolumn{1}{c}{\multirow{1}*{Price}}
		&\multicolumn{1}{c}{0.1036$\pm$0.0182}
		&\multicolumn{1}{c}{0.1031$\pm$0.0168}
		&\multicolumn{1}{c}{0.1006$\pm$0.0142}
		&\multicolumn{1}{c}{\textbf{0.0985}$\pm$\textbf{0.0137}}
		&\multicolumn{1}{c}{0.0997$\pm$0.0161}\\
	
		\multicolumn{1}{c}{\multirow{1}*{Machine-CPU}}
		&\multicolumn{1}{c}{0.0646$\pm$0.0260}
		&\multicolumn{1}{c}{0.0573$\pm$0.0182}
		&\multicolumn{1}{c}{0.0544$\pm$0.0156}
		&\multicolumn{1}{c}{\textbf{0.0530}$\pm$\textbf{0.0163}}
		&\multicolumn{1}{c}{0.0534$\pm$0.0164}\\
	
		\bottomrule
	\end{tabular}
\end{table*}

\begin{table*}[]\footnotesize
	\renewcommand\arraystretch{1.5}
	\setlength{\abovecaptionskip}{0pt}
	\setlength{\belowcaptionskip}{5pt}
	\centering
	\caption{Training time(sec) of five algorithms}
	\begin{tabular}{m{0.1cm}m{0.1cm}m{0.1cm}m{0.1cm}m{0.1cm}m{0.1cm}}
		\toprule
		\multicolumn{1}{c}{\multirow{1}*{Datasets}} & \multicolumn{1}{c}{ELM} & \multicolumn{1}{c}{RELM}& \multicolumn{1}{c}{ELM-RCC}& \multicolumn{1}{c}{ELM-MEE}& \multicolumn{1}{c}{ELM-QMEE}\\
		%\cline{2-6}
		%&\multicolumn{1}{c}{Tr-time}&\multicolumn{1}{c}{Tr-time}&\multicolumn{1}{c}{Tr-time}&\multicolumn{1}{c}{Tr-time}&\multicolumn{1}{c}{Tr-time} \\
		\hline
		\multicolumn{1}{c}{Servo}	&\multicolumn{1}{c}{0.0020$\pm$0.0082}	&\multicolumn{1}{c}{0.0011$\pm$0.0040}	&\multicolumn{1}{c}{0.0127$\pm$0.0184}	&\multicolumn{1}{c}{1.0286$\pm$0.0116}	&\multicolumn{1}{c}{0.0314$\pm$0.0181}	\\
		\multicolumn{1}{c}{Yacht}	&\multicolumn{1}{c}{0.0056$\pm$0.0125}	&\multicolumn{1}{c}{0.0048$\pm$0.0103}	&\multicolumn{1}{c}{0.0641$\pm$0.0325}	&\multicolumn{1}{c}{59.9422$\pm$2.1326}	&\multicolumn{1}{c}{0.1086$\pm$0.0340}	\\
		\multicolumn{1}{c}{Computer Hardware}	&\multicolumn{1}{c}{0.0022$\pm$0.0067}	&\multicolumn{1}{c}{0.0014$\pm$0.0067}	&\multicolumn{1}{c}{0.0050$\pm$0.0125}	&\multicolumn{1}{c}{4.7689$\pm$0.2338}	&\multicolumn{1}{c}{0.0716$\pm$0.0271}	\\
		\multicolumn{1}{c}{Price}	&\multicolumn{1}{c}{$1.5625 \times 10^{-4}$$\pm$0.0016}	&\multicolumn{1}{c}{0.0651$\pm$0.0086}	&\multicolumn{1}{c}{$7.8125 \times 10^{-4}$$\pm$0.0034}	&\multicolumn{1}{c}{0.5859$\pm$0.0093}	&\multicolumn{1}{c}{0.0233$\pm$0.0129}	\\
		\multicolumn{1}{c}{Machine-CPU}	&\multicolumn{1}{c}{0.0011$\pm$0.0056}	&\multicolumn{1}{c}{$3.1250 \times 10^{-4}$$\pm$0.0022}	&\multicolumn{1}{c}{0.0027$\pm$0.0063}	&\multicolumn{1}{c}{1.0913$\pm$0.0121}	&\multicolumn{1}{c}{0.0223$\pm$0.0114}	\\
		\bottomrule
	\end{tabular}
\end{table*}

\subsection{Echo State Networks}

In the last example, we apply the QMEE to train an echo state network (ESN) \cite{jaeger2001echo,jaeger2004harnessing,lukovsevivcius2009reservoir}, a new paradigm in recurrent neural network (RNN)\cite{mandic2001recurrent,lee2000identification}. The ESN randomly builds a large sparse reservoir to replace the hidden layer of RNN, which overcomes the shortcomings of complicated computation and difficulties in determining the network topology of a traditional RNN.

We consider a discrete-time ESN with $P$ input units, $L$ internal network units and $Q$ output units. The dynamic and output equations of the standard ESN can be written as follows:

\begin{equation}\label{esn}
\left\{ \begin{array}{l}
x\left( {k + 1} \right) = f\left( {{\textbf{W}^x}x\left( k \right) + {\textbf{W}^{in}}u\left( {k + 1} \right) + {\textbf{W}^{fb}}y\left( k \right)} \right)\\
y\left( k \right) = g\left( {{\textbf{W}^{out}}\varphi \left( k \right) } \right)
\end{array} \right.
\end{equation}

\noindent where $\varphi \left( k \right) = \left( \begin{array}{l}
u\left( k \right)\\
x\left( k \right)
\end{array} \right)$, $f = \left( {{f_1} \ldots {f_L}} \right)$ is the nonlinear activation function of reservoir units, $g = \left( {{g_1} \ldots {g_{\rm{Q}}}} \right)$ is a linear or nonlinear activation function of the output layer, ${\textbf{W}^{in}}$ is an $L \times P$ input weight matrix, ${\textbf{W}^{x}}$ is an $L \times L$ internal connection weight matrix of the reservoir, ${\textbf{W}^{fb}}$ is an $L \times Q$ weight matrix that feeds back the output to the reservoir units, and ${\textbf{W}^{out}}$ is an $Q \times (P+L)$ output weight matrix. To establish an ESN described above, with the property of echo states, the weight matrix ${\textbf{W}^{x}}$ must satisfy the condition ${\sigma _{\max }} < 1$, with ${\sigma _{\max }} $ being the largest singular value of ${\textbf{W}^{x}}$. In this article we assume that ${\textbf{W}^{fb}}=0$. The weight matrices ${\textbf{W}^{in}}$ and ${\textbf{W}^{x}}$ are randomly determined. Then the nonlinear system can be converted to:

\begin{equation}
\textbf{Y} = {\textbf{W}^{out}}\textbf{X}
\end{equation}

\noindent where the $k$th column of the matrix $\textbf{X}$ is $ \varphi \left( k \right) $ . The optimal solution of ${\textbf{W}^{out}}$ under MSE criterion can be obtained by ${\textbf{W}^{out}} = {\left( {{\textbf{X}^T}\textbf{X}} \right)^{ - 1}}{\textbf{X}^T}\textbf{Y}$. Here, we use the following QMEE cost function to train the ESN:

\begin{equation}
{J_{QMEE}}\left( {{\textbf{W}^{out}}} \right)\; = \frac{1}{{{N^2}}}\sum\limits_{i = 1}^N {\left( {\sum\limits_{m = 1}^M {{M_m}{G_\sigma }\left( {{e_i} - {c_m}} \right)} } \right)} 
\end{equation}

\noindent where ${\textbf{e}_i} = {\textbf{t}_i} - {\textbf{y}_i}$, with $\textbf{t}_i$ and $\textbf{y}_i$ being respectively, the $i$th rows of the target matrix $\textbf{T}$ and output matrix $\textbf{Y}$. Different approaches can be used to solve the above optimization problem. Here, the Root Mean Square Propagation (RMSProp) is used. The RMSProp as a variant of stochastic gradient descent (SGD) has been widely used in deep learning. With RMSProp the output weights can be updated by

\begin{equation}\label{up_w}
w_i^{out}\left( {k + 1} \right) = w_i^{out}\left( k \right) - \frac{\eta }{{\sqrt {v + r} }}{\nabla _{w_i^{out}}}{J_{QMEE}}\left( k \right)
\end{equation}

\begin{equation}\label{up_v}
v = \rho v + \left( {1 - \rho } \right){\left( {{\nabla _{w_i^{out}}}{J_{QMEE}}\left( k \right)} \right)^2}
\end{equation}

\noindent where $\eta $ is the learning rate parameter, $r$ is a small positive constant and $\rho $ is the forgetting factor. The gradient term ${\nabla _{w_i^{out}}}{J_{QMEE}}\left( n \right)$ can be computed as 

\begin{equation}\label{gradient_term}
\begin{aligned}
&{\nabla _{w_i^{out}}}{J_{QMEE}}\left( k \right) \!=\! \sum\limits_{m = 1}^M {M_m}\exp \!\left(\! { \!-\! \frac{{{{\left( {{t_i}\left( k \right) \!-\! w_i^{out}\left( k \right)x\left( k \right) \!-\! {c_m}} \right)}^2}}}{{2{\sigma ^2}}}} \!\!\right)\\
&\left( {{t_i}\left( k \right) - w_i^{out}\left( k \right)x\left( k \right) - {c_m}} \right)x'\left( k \right) 
\end{aligned}
\end{equation}

\noindent The learning algorithm of the ESN under QMEE is given in \textit{Algorithm 4}, called ESN-QMEE in this paper.

\begin{algorithm}
	\renewcommand{\algorithmicrequire}{\textbf{Input:}}
	\renewcommand{\algorithmicensure}{\textbf{Output:}}
	\caption{ESN-QMEE}
	\label{alg:1}
	\begin{algorithmic}[1]
		\REQUIRE samples $\{ {\textbf{u}_i},{\textbf{t}_i}\} _{i = 1}^N$
		\ENSURE weight matrix ${\textbf{W}^{out}}$ 
		\STATE Parameters setting: learning rate $\eta$, constant $r$, forgetting factor $\rho$, iteration number $K$, kernel width $\sigma$, quantization threshold $\epsilon$    
		\STATE Initialization: set the number and the sparseness of reservoir units, randomly initialize ${\textbf{W}^{x}}$ and ${\textbf{W}^{in}}$ , and compute the matrix $\textbf{X}$ 
		\FOR{$ k = 2,...,K $}
		\STATE Compute the gradient term ${\nabla _{w_i^{out}}}{J_{QMEE}}\left( k \right)$ by (\ref{gradient_term})
		\STATE Compute the term $v$ by (\ref{up_v})
		\STATE Update the weight matrix ${\textbf{W}^{out}}$ by (\ref{up_w})
		\ENDFOR
	\end{algorithmic}  
\end{algorithm}

Next, we apply the proposed ESN-QMEE to the short-term prediction of the Mackey-Glass (MG) chaotic time series, compared to some other ESN algorithms. The MG dynamic system is governed by the following time-delay ordinary differential equation \cite{glass1979pathological}

\begin{equation}
\frac{{dx}}{{dt}} = ax\left( t \right) + \frac{{bx\left( {t - \tau } \right)}}{{1 + x{{\left( {t - \tau } \right)}^{10}}}}
\end{equation}

\noindent with $a=-0.1$, $b=0.2$, $\tau=17$. This system has a chaotic attractor if $\tau>16.8$ . In this article, we choose the delay time and the embedded dimension as six and four, which are determined by the mutual information \cite{fraser1986independent}, i.e. the vector ${\left[ {x\left( {t - 24} \right),x\left( {t - 18} \right),x\left( {t - 12} \right),x\left( {t - 6} \right)} \right]^T}$ is used as the input to predict the present value $x(t)$ that is the desired response in this example. In the simulation, the number of reservoir units is set to 400. The spectral radius and the sparseness of ${\textbf{W}^x}$ are 0.95 and 0.01. A segment of 900 samples are used as the training data and another 400 samples as the testing data. The noise model ${v_i} = \left( {1 - {a_i}} \right){A_i} + {a_i}{B_i}$ mentioned in the subsection A is used to generate the noises added to the training data, where the occurrence probability is $c=0.2$, $B_i$ is a white Gaussian process with zero-mean and variance 0.01, and $A_i$ is a mixture Gaussian process with density $0.5\mathcal{N}\left( {\alpha ,0.01} \right) + 0.5\mathcal{N}\left( { - \alpha ,0.01} \right)$. Further, the normalized root mean squared error (NRMSE) is used to measure the performance of different algorithms, given by

\begin{equation}\label{nrmse}
NRMSE = \sqrt {\frac{1}{{N\sigma _{target}^2}}\sum\limits_{n = 0}^{N - 1} {{{\left( {t\left( n \right) - y\left( n \right)} \right)}^2}} } 
\end{equation}

\noindent which $\sigma _{target}^2$ denotes the variance of the target signal. Similar to the previous example, the NRMSE of the MEE and QMEE will be calculated by adding a bias value to the testing errors. The parameter settings of five ESN algorithms are given in Table VII. The NRMSEs of six ESN algorithms over 10 Monte Carlo runs for different values of $\alpha$ are illustrated in Fig. 3, and the corresponding training times are shown in Table VIII. Once again, the QMEE based algorithm can outperform other algorithms, whose performance is very close to that of the MEE based algorithm but with much less computational cost.

\begin{table}[]\footnotesize
	\renewcommand\arraystretch{1.5}
	\setlength{\abovecaptionskip}{0pt}
	\setlength{\belowcaptionskip}{5pt}
	\centering
	\caption{Parameter settings of five ESN algorithms in chaotic time series prediction}
	\begin{tabular}{ccccccc}
		\toprule
		\multirow{2}*{$\alpha$} & RESN&LESN\cite{jaeger2007optimization}&CESN\cite{guo2017robust}&ESN-MEE & \multicolumn{2}{c}{ESN-QMEE}\\
		\cline{2-7}
		&$\lambda$&L&$\sigma$&$\sigma$&$\sigma$&$\gamma $\\
		\hline
		0.1&0.01&0.94&6.3&0.06&0.8&0.07\\
		0.2&0.01&0.92&4.0&0.07&0.7&0.01\\
		0.3&0.1&0.93&3.0&0.08&0.7&0.02\\
		0.4&0.1&0.93&3.0&0.08&0.7&0.03\\
		\bottomrule
	\end{tabular}
\end{table}

\begin{figure}[htbp]
	\setlength{\abovecaptionskip}{0pt}
	\setlength{\belowcaptionskip}{0pt}
	\centering
	\includegraphics[width=\linewidth]{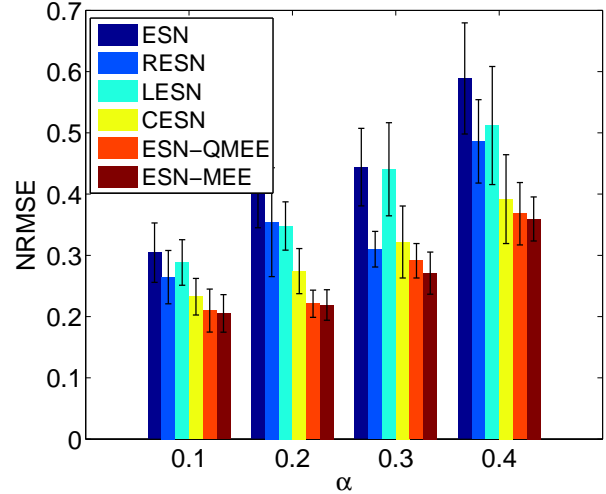}
	\caption{NRMSE with different values of $\alpha$}
	\label{fig1}
\end{figure}

\begin{table*}[]\footnotesize
	\renewcommand\arraystretch{1.5}
	\setlength{\abovecaptionskip}{0pt}
	\setlength{\belowcaptionskip}{5pt}
	\centering
	\caption{Parameter settings of five ESN algorithms in chaotic time series prediction}
	\begin{tabular}{m{0.8cm}cccccc}
		\toprule
		&$\quad$ESN& RESN&LESN&CESN&ESN-MEE & ESN-QMEE\\
		
		\hline
		Training time (Sec)&$ 0.0336 \pm 0.0029 $ &	$0.0330 \pm 5.5606 \!\!\times\!\! 10^{-4} $&	$ 0.0328\pm0.0010 $&	$ 1.4505\pm0.1640 $&	$ 2.9903\!\!\times\!\! 10^{3} \pm10.3857 $&	$ 58.3307\pm1.1894 $\\
		
		\bottomrule
	\end{tabular}
\end{table*}

\section{CONCLUSION}
Minimum error entropy (MEE) criterion can outperform traditional MSE criterion in non-Gaussian signal processing and machine learning. However, it is computationally much more expensive due to the double summation operation in the objective function, resulting in computational expense scaling as $O(N^2)$, where $N$ is the number of samples. In this paper, we proposed a simplified MEE criterion, called quantized MEE (QMEE), whose computational complexity is $O(MN)$, with $M \ll N$. The basic idea is to reduce the number of the inner summations by quantizing the error samples. Some important properties of the QMEE are presented. Experimental results with linear and nonlinear models (such as ELM and ESN) confirm that the proposed QMEE can achieve almost the same performance as the original MEE criterion, but needs much less computational time.

\bibliographystyle{unsrt}
\bibliography{QMEE}
\end{document}